\documentclass[11pt,a4paper]{article}
\usepackage[hyperref]{emnlp2020}
\usepackage{times}
\usepackage{latexsym}

\usepackage{graphicx}

\usepackage{microtype}

\usepackage{booktabs}

\aclfinalcopy 
\def\aclpaperid{1923}

\newcommand\taska{\textsc{Base}}
\newcommand\taskb{\textsc{Paragraph}}
\newcommand\taskc{\textsc{EditPremise}}
\newcommand\taskd{\textsc{EditOther}}
\newcommand\taskf{\textsc{Contrast}} 

\newcommand\mnli{\textsc{MNLI}}
\newcommand\anli{\textsc{ANLI}}

\title{New Protocols and Negative Results\\ for Textual Entailment Data Collection}

\author{Samuel R. Bowman\thanks{\,\,\,Work done while visiting Google.} \\
  Center for Data Science, Department of Linguistics, and Department of Computer Science\\New York University\\
  {\tt bowman@nyu.edu} \AND
  Jennimaria Palomaki, Livio Baldini Soares, Emily Pitler\\
  Google Research \\
  {\tt \{jpalomaki,liviobs,epitler\}@google.com}
}

\date{}

\begin{document}
\maketitle
\begin{abstract}
Natural language inference (NLI) data has proven useful in benchmarking and, especially, as pretraining data for tasks requiring language understanding. However, the crowdsourcing protocol that was used to collect this data has known issues and was not explicitly optimized for either of these purposes, so it is likely far from ideal. We propose four alternative protocols, each aimed at improving either the ease with which annotators can produce sound training examples or the quality and diversity of those examples. Using these alternatives and a fifth baseline protocol, we collect and compare five new 8.5k-example training sets. In evaluations focused on transfer learning applications, our results are solidly negative, with models trained on our baseline dataset yielding good transfer performance to downstream tasks, but none of our four new methods (nor the recent ANLI) showing any improvements over that baseline. In a small silver lining, we observe that all four new protocols, especially those where annotators edit \textit{pre-filled} text boxes, reduce previously observed issues with annotation artifacts.

\end{abstract}

\section{Introduction} 

\begin{figure}[t!]
\centering
\includegraphics[width=.95\columnwidth]{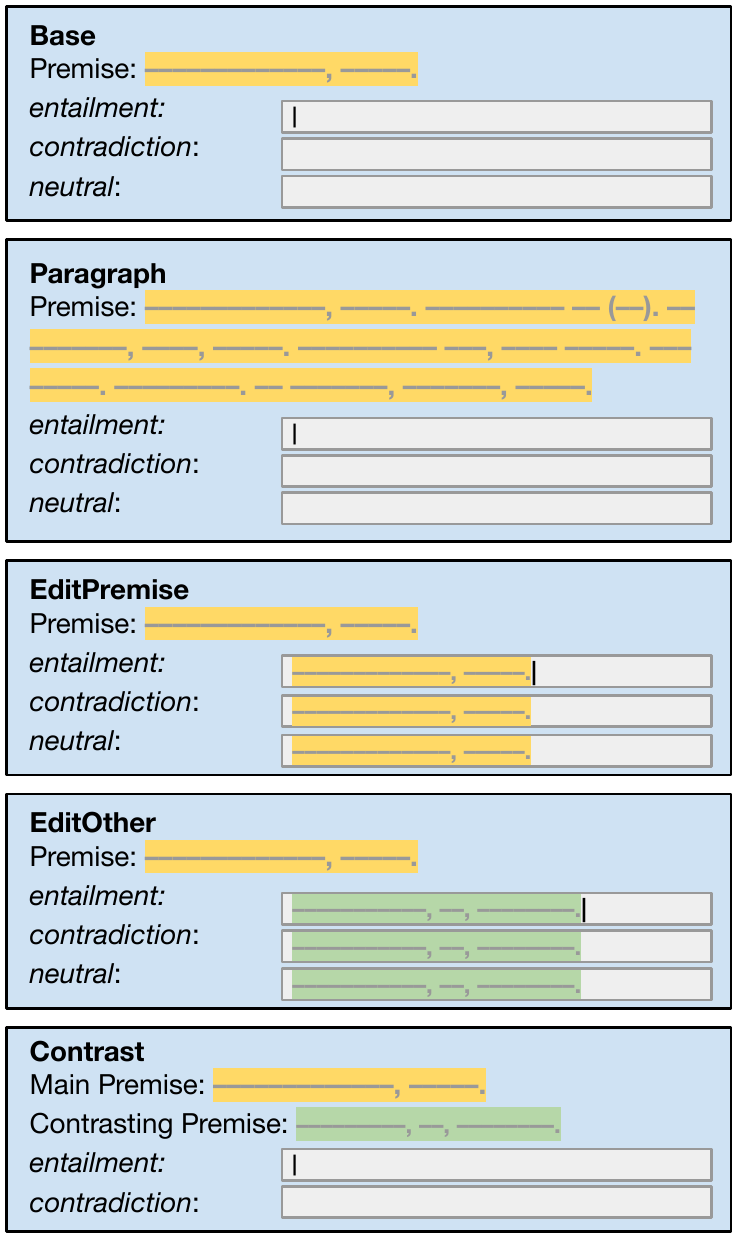}
\caption{The annotation interfaces we evaluate.}
\label{fig:interface}
\end{figure}

The task of natural language inference (NLI; also known as textual entailment) has been widely used as an evaluation task when developing new methods for language understanding tasks, and it has recently become clear that high-quality NLI data can be useful in transfer learning as well, driving much of the recent use of the task: Several recent papers have shown that training large neural network models on natural language inference data, then fine-tuning them for other language understanding tasks often yields substantially better results on those target tasks \citep{conneau2017supervised,subramanian2018learning}. This result holds even when starting from large models like BERT \citep{devlin2018bert} that have \textit{already} been pretrained extensively on unlabeled data \citep{phang2018sentence,clark2019boolq,liu2019roberta,wang2019superglue}.
    
The largest general-purpose corpus for NLI, and the one that has proven most successful in this setting, is the Multi-Genre NLI Corpus \citep[MNLI][]{williams2018broad}. MNLI was designed informally for use in a benchmark task (with no consideration of transfer learning), and in any case,
no explicit experimental research went into its design. Further, data collected under MNLI's data collection protocol 
has known issues with \textit{annotation artifacts} which make it possible to perform much better than chance using only one of the two sentences that make up each example \citep{TSUCHIYA18.786,gururangan2018,poliak-2018-dnc}.

This work experimentally evaluates four potential changes to the original MNLI data collection protocol that are designed to improve either the ease with which annotators can produce sound training examples or the quality and diversity of those examples. We collect a baseline dataset of 8.5k examples that follows the MNLI protocol with our annotator pool, followed by four additional datasets of the same size which isolate each of our candidate changes. (See Figure \ref{fig:interface} for a schematic.) We then compare all five in a set of experiments, focused on transfer learning, that look at our ability to use each of these datasets to improve performance on the eight downstream language understanding tasks in the SuperGLUE \citep{wang2019superglue} benchmark.

All five of our datasets are consistent with the task definition that was used in MNLI, which is in turn based on the definition introduced by \citet{dagan2006pascal}. In this task, each example consists of a pair of short texts: a \textit{premise} and a \textit{hypothesis}. The model is asked to read both texts and make a three-way classification decision: Given the premise, would a reasonable person infer that hypothesis must be true (\textit{entailment}), that it must be false (\textit{contradiction}), or that there is not enough information to judge (\textit{neutral})?
While it is certainly not clear that this design is optimal for any application, we leave a more broad-based exploration of task definitions for future work.

Our \textbf{\taska} data collection protocol follows MNLI closely in asking  annotators to read a premise sentence and then write three corresponding hypothesis sentences in empty text boxes corresponding to the three different labels (\textit{entailment}, \textit{contradiction}, and \textit{neutral}). When an annotator follows this protocol, they produce three sentence pairs at once, all sharing a single premise.

Our \textbf{\taskb} protocol tests the effect of supplying annotators with complete paragraphs, rather than sentences, as premises. Longer texts offer  the potential for discourse-level inferences, the addition of which should yield a dataset that is more difficult, more diverse, and less likely to contain trivial artifacts. However, one might expect that asking annotators to read full paragraphs should increase the time required to create a single example; time which could potentially be better spent creating \textit{more} examples.

Our \textbf{\taskc} and \textbf{\taskd} protocols test the effect of pre-filling a single seed text in each of the three text boxes that annotators are asked to fill out. By reducing the raw amount of typing required, this could allow annotators to produce good examples more quickly. By encouraging them to keep the three sentences similar, it could also indirectly facilitate the construction of minimal-pair-like examples that minimize artifacts, in the style of \citet{kaushik2019learning}. We test two variants of this idea: One uses a copy of the premise sentence as a seed text and the second retrieves a new sentence from an existing corpus that is similar to the premise sentence, and uses that.

Our \textbf{\taskf} protocol tests the effect of adding artificial constraints on the kinds of hypothesis sentences annotators can write. Giving annotators difficult and varying constraints could encourage creativity and prevent annotators from falling into patterns in their writing that lead to easier or more repetitive data. However, as with the use of longer contexts in \taskb, this protocol risks substantially slowing the annotation process. We experiment with a procedure inspired by that used to create the language-and-vision dataset NLVR2 \citep{suhr-etal-2019-corpus},  in which annotators must write sentences that show some specified relationship (entailment or contradiction) to a given premise, but do \textit{not} show that relationship to a second similar distractor premise.

Because we see transfer learning as the primary application area for which it would be valuable to collect additional large-scale NLI datasets, we focus our evaluation on this setting, and do not collect or designate test sets for the experimental datasets we collect.
In transfer evaluations on the SuperGLUE benchmark \citep{wang2019superglue}, our \taska{} dataset and the datasets collected under our four new protocols offer substantial improvements in transfer ability over a plain RoBERTa or XLNet model, comparable to the gains seen with an equally-sized sample of MNLI. However, \taska{} reliably shows the strongest transfer results. This finding, combined with a low variance across runs, strongly suggests that none of these four interventions improves the suitability of NLI data for transfer learning. 
We also observe that \taska{}, \taskb{}, \taskc{}, and \taskd{} all require very similar amounts of annotator time, reducing the potential downside of \taskb{}, but also invalidating the primary motivation behind \taskc{} and \taskd{}. 
While our primary results are negative, we also observe that all four of these methods produce data of comparable subjective quality to \taska{} while significantly reducing the incidence of previously reported annotation artifacts.

\section{Related Work}

\begin{table}[t!] 

\centering\small
\begin{tabular}{p{.945\columnwidth}}
\toprule
\textbf{\mnli\ (Training Set)}\\
\midrule
\textbf{P:} \textit{Conceptually cream skimming has two basic dimensions - product and geography.}\\ 
\textbf{H:} \textit{Product and geography are what make cream skimming work.}\\
\textbf{neutral}\\

\midrule
\textbf{\taska}\\
\midrule
\textbf{P:} \textit{The board had also expressed concerns about the amounts of cash kept by SNC's Libyan office, at that time approximately \$10 million, according to the company's chief financial officer.}	\\

\textbf{H:} \textit{According to the board, the Libyan office should be holding more cash on hand.}\\

\textbf{contradiction} \\

\midrule
\textbf{\taskb}\\
\midrule
\textbf{P:} \textit{The paper, along with the ''Washington Blade'', was acquired by Window Media, LLC in 2001, and both were then sold to HX Media in 2007. Kat Long succeeded Trenton Straube as editor-in-chief in February 2009. The paper ceased publication in July 2009.} \\

\textbf{H:} \textit{Kal Long succeeded Trenton Straube as editor-in-chief in March 2019.} \\

\textbf{contradiction}\\

\midrule
\textbf{\taskc}\\
\midrule
\textbf{P:} \textit{This standpoint is believed to promote Deaf people's right to collective space within society to pass on their language and culture to future generations.} \\

\textbf{H:} \textit{This standpoint is believed to demote Deaf people's right to collective space within society.} \\

\textbf{contradiction} \\

\midrule
\textbf{\taskd}\\
\midrule
\textbf{P:} \textit{Shobhona Sharma (born 5 February 1953) is a professor specializing in immunology, molecular biology, and biochemistry at the Tata Institute of Fundamental Research, Mumbai.}\\

\textbf{H:} \textit{Shobhona Sharma is also professor of mathematics at the Tata Institute of Fundamental Research, Mumbai.}\\

\textbf{neutral}\\
\midrule
\textbf{\taskf}\\
\midrule
\textbf{P:} \textit{Bengt Erik Johan Renvall (September 22, 1959 -- August 24, 2015) was a Swedish dancer and choreographer active in the United States from 1978.}\\

\textbf{H:} \textit{He was a dancer in America in the 1970s.}\\

\textbf{entailment}\\
\bottomrule
\end{tabular}

\caption{Randomly selected examples from the datasets under study. Neither the MNLI training set nor any of our collected data are filtered for quality in any way, and errors or debatable judgments are common in both.}\label{tab:ex}

\end{table}

Existing NLI datasets have been built using a wide range of strategies: FraCaS \citep{cooper1996} and several targeted  evaluation sets were constructed manually by experts from scratch. The RTE challenge corpora \citep[][et seq.]{dagan2006pascal} primarily used expert annotations on top of existing premise sentences. SICK \citep{marelli2014sick} was created using a structured pipeline centered on asking crowdworkers to edit sentences in prescribed ways. MPE \citep{lai-etal-2017-natural} uses a similar strategy, but constructs unordered \textit{sets} of sentences for use as premises.
SNLI \citep{snli:emnlp2015} introduced the method, also used in MNLI, of asking crowdworkers to compose labeled hypotheses for a given premise. SciTail \citep{scitail} and SWAG \citep{zellers2018swag} used domain-specific resources to pair up existing sentences as potential entailment pairs for annotation, with SWAG additionally using trained models to select the examples most worth annotating. There has been little work directly evaluating and comparing these many methods. In that absence, we focus on the SNLI/MNLI approach, because it has been shown to be effective for the collection of pretraining data and because its reliance on only crowdworkers and unstructured source text makes it simple 
to scale.

Two recent papers have investigated methods that could augment the base MNLI protocol we study here. ANLI \citep{nie2019adversarial} collects new examples following this protocol, but adds an incentive for crowdworkers to produce sentence pairs on which a baseline system will perform poorly. \citet{kaushik2019learning} introduce a method for expanding an already-collected dataset by making minimal edits to existing examples that change their labels, with the intent to better teach models to isolate the factors that are causally responsible for the label assignments. Both of these papers offer methodological changes that are potentially complementary to the changes we investigate here, and neither evaluates the impact of their methods on transfer learning. Since ANLI is large and roughly comparable with MNLI, we include it in our transfer evaluations here.

In addition to NLVR2 (which motivated our \taskf{} protocol), WinoGrande \citep{sakaguchi2019winogrande} also showed promising results from the use of artificial constraints during the annotation process for another style of dataset.

The observation that NLI data can be effective in pretraining was first reported for SNLI and MNLI by \citet{conneau2017supervised} on models pretrained from scratch on NLI data. This finding was replicated in the setting of multi-task pretraining by \citet{subramanian2018learning}. This was later extended to the context of \textit{intermediate training}---where a model is pretrained on unlabeled data, then on relatively abundant labeled data (MNLI), and finally scarce task-specific labeled data---by \citet{phang2018sentence}, \citet{clark2019boolq}, \citet{liu2019mt}, \citet{yang2019xlnet}, and \citet{liu2019roberta} across a range of large pretrained models models and target language-understanding tasks. Similar results have been observed with transfer from the other reasoning-oriented datasets \citep{sap2019socialiqa,bhagavatula2019abductive}, especially to target tasks centered on common sense.
Another related body of work  \citep{mou-etal-2016-transferable,bingel-sogaard:2017:EACLshort,wang-etal-2019-tell,pruksachatkun2020} has explored the broader empirical landscape of which supervised NLP tasks can offer effective pretraining for other supervised NLP tasks.

\section{Data Collection}

The annotation interface for our tasks is similar to that used for SNLI and MNLI: We provide a premise from a preexisting text source and ask human annotators to provide three hypothesis sentences: one that says something true about the fact or situation in the prompt (\textit{entailment}), one that says something that may or may not be true about the fact or situation in the prompt (\textit{neutral})---with the additional instruction that this sentence should discuss the same topic as the prompt but could be either true or false because the prompt does not provide enough information to be sure---and one that definitely does not say something true about the fact or situation in the prompt (\textit{contradiction}). 

We evaluate five variants of this interface:

\paragraph{\taska}

We show annotators a premise sentence and ask them to compose one new sentence for each label.

\paragraph{\taskb}

We use the same instructions as \taska, but with full paragraphs, rather than single sentences, as the supplied premises. 

\paragraph{\taskc}

We pre-fill three text boxes with editable copies of the premise sentence, and ask annotators to edit each text field to compose sentences that match the three different labels. Annotators may delete the pre-filled text.

\paragraph{\taskd} 

We follow the same procedure as \taskc, but rather than pre-filling the premise as a seed sentence, we instead use a similarity search method to retrieve a new sentence that is similar to the premise.

\paragraph{\taskf}

We again retrieve a second sentence that is similar to the premise, but we display it as a \textit{contrasting premise} rather than using it to seed an editable text box. We then ask annotators to compose two new sentences: One sentence must be true \textit{only} about the fact or situation in the first premise (that is, contradictory or neutral with respect to the contrasting premise). The other sentence must be \textit{false} only about the fact or situation in the first premise (and \textit{true} or neutral with respect to the contrasting premise). This yields an entailment pair and a contradiction pair, both of which use only the first  premises, with the contrasting premise serving only as a constraint on the annotation process. We could not find a sufficiently intuitive way to collect neutral sentence pairs under this protocol and opted to use only two classes rather than increase the difficulty of an already unintuitive task.

\subsection{Text Source}

MNLI uses the small but stylistically diverse OpenANC corpus \citep{ide2006openANC} as its source for premise sentences, but uses nearly every available sentence from its non-technical sections, making it impractical for our use. To avoid  re-using premise sentences,
We instead draw on English Wikipedia.\footnote{We use the 2019-06-20 downloadable version, remove markup and tables with Apertium's  WikiExtractor feature \citep{forcada2011apertium}, sentence-tokenize it with SpaCy \citep{spacy2}, and randomly sample sentences (or paragraphs) for annotation.}

\paragraph{Similarity Search}

The \taskd\ and \taskf\ protocols require pairs of similar sentences as their inputs. To construct these, we assemble a heuristic sentence-matching system intended to generate pairs of highly similar sentences that can be minimally edited to construct entailments or contradictions: 
Given a premise, we retrieve its closest 10k nearest neighbors according to dot-product similarity over Universal Sentence Encoder~\citep{cer-etal-2018-universal} embeddings. Using a parser and an NER system, we then select those neighbors which share a subject noun phrase in common with the premise (dropping premises for which no such neighbors exist). From those filtered neighbors, we retrieve the single non-identical neighbor that has the highest overlap with the premise in both raw tokens and entity mentions, preferring sentences with similar length to the hypothesis.\footnote{For dependency parse and named entity recognition annotations, we use the Google Natural Language API: \texttt{\url{https://cloud.google.com/natural-language/}}.}

\begin{table}[t]
\small

\centering
\begin{tabular}{llcc}
\toprule
& \textbf{Label} & \textbf{Length} & \textbf{Unique} \\
  &                & \textbf{$\mu$ $(\sigma)$}       & \textbf{$\mu$ $(\sigma)$} \\
\midrule
\midrule
 \multicolumn{4}{c}{\textbf{MNLI 8.5k}}      \\
 \midrule
 \textbf{premise}    & all labels    &  23.2 (15.8) & --- \\ 
  \textbf{hypothesis} & entailment    &  11.4	(4.8) &	4.5	(4.8) \\
 \textbf{hypothesis} & neutral       &  12.5 (4.8) & 7.2 (4.8) \\
 \textbf{hypothesis} & contradiction &  11.0 (4.3) &  5.8 (4.3) \\
 \midrule

\multicolumn{4}{c}{\textbf{MNLI Gov. 8.5k}}      \\
\midrule
\textbf{premise}    & all labels    & 25.1 (13.4) &  --- \\ 
\textbf{hypothesis} & entailment    & 12.6 (5.1)  & 4.4 (5.1) \\
\textbf{hypothesis} & neutral       & 13.0 (5.3)  &	7.1 (5.3) \\
\textbf{hypothesis} & contradiction & 12.0 (4.5)  & 5.7	(4.5) \\
\midrule
\multicolumn{4}{c}{\textbf{\taska}}      \\
\midrule
\textbf{premise}    & all labels & 23.3 (11.4) &   --- \\ 
\textbf{hypothesis} & entailment       & 10.6 (5.6) & 2.3 (5.5) \\
\textbf{hypothesis} & neutral       & 10.5 (5.5) & 4.5 (5.5) \\
\textbf{hypothesis} & contradiction & 10.2 (5.1) & 4.0 (5.1) \\
\midrule
\multicolumn{4}{c}{\textbf{\taskb}}       \\
\midrule
\textbf{premise}    & all labels    & 66.7 (60.0) &  --- \\ 
\textbf{hypothesis} & entailment    & 13.0 (8.1) & 2.3	(8.1) \\
\textbf{hypothesis} & neutral       & 12.9 (8.1) & 4.1 (8.1) \\
\textbf{hypothesis} & contradiction & 12.5 (7.9) & 3.3 (7.9)  \\
\midrule
\multicolumn{4}{c}{\textbf{\taskc}}       \\
\midrule
\textbf{hypothesis} & entailment    & 15.0 (8.9) & 2.5 (8.9) \\
\textbf{hypothesis} & neutral       & 17.0 (9.8) & 4.3 (9.8) \\
\textbf{hypothesis} & contradiction & 15.3 (9.2) & 3.3 (9.2) \\
\midrule
\multicolumn{4}{c}{\textbf{\taskd}}       \\
\midrule
\textbf{hypothesis} & entailment       & 12.6 (6.3) & 3.2 (6.3) \\
\textbf{hypothesis} & neutral       & 13.0 (6.8) & 6.2 (6.8) \\
\textbf{hypothesis} & contradiction & 12.7 (6.4) & 4.7 (6.4) \\
\midrule
\multicolumn{4}{c}{\textbf{\taskf}}     \\
\midrule
\textbf{hypothesis} & entailment    & 7.9 (5.1) & 2.5 (5.1)  \\
\textbf{hypothesis} & contradiction & 7.7 (4.9) & 3.5 (4.9) \\ 
\bottomrule
\end{tabular}
\caption{Key text statistics. 
Premises are drawn from essentially the same distribution in all our tasks except \taskb, so are shown only once.
The \textit{Unique} column shows the number of tokens that appear in a hypothesis but not in the corresponding premise.}\label{tab:length}
\end{table}

\subsection{The Annotation Process}

We start data collection for each protocol with a pilot of 100 items, which are not included in the final datasets. We use these to refine task instructions and to provide feedback to our annotator pool on the intended task definition. We continue to provide regular feedback throughout the annotation process to clarify ambiguities in the protocols and to discourage the use of systematic patterns---such as consistently composing shorter hypotheses for entailments than for contradictions---that could make the resulting data artificially easy. 

Annotators are allowed to skip prompts which they deem unusable for any reason. These generally involve either non-sentence strings that were mishandled by our sentence tokenizer or premises with inaccessible technical language. Skip rates ranged from about 2.5\% for \taskd\ to about 10\% for \taskf\ (which can only be completed when the two premises are both comprehensible and sufficiently different from one another).

A pool of 19 professional annotators located in the United States worked on our tasks, with about ten working on each protocol.
As a consequence of this relatively small annotation team, many annotators worked under more than one protocol, which we ran consecutively. This introduces a modest potential bias against \taska{}, in that annotators start the later tasks having seen somewhat more feedback. 

Because of our focus on collecting training data for transfer learning applications, we do not use any kind of second-pass annotation process for quality control, and we neither designate a test set nor recommend the use of our released datasets for system evaluation. We aim to use our limited annotation time budget to collect the largest and best possible sample of (pre)training data, and we are motivated by work like \citet{khetan2017learning} which calls into question the value of second-pass quality-control annotations for training data.

\subsection{The Resulting Data}

Using each protocol, we collect a training set of exactly 8,500 examples and a small validation set of at least 300 examples.\footnote{available for download at \url{https://github.com/google-research-datasets/Textual-Entailment-New-Protocols}} 
Table \ref{tab:ex} shows examples.

Hypotheses are mostly fluent, full sentences that adhere to 
writing conventions for US English. In constructing  hypotheses, annotators often reuse words or phrases from the premise, but rearrange them, alter their inflectional forms, or substitute synonyms or antonyms. Hypotheses tend to differ 
from premises both grammatically and stylistically. 

Table \ref{tab:length} shows some statistics for the collected text. The two methods that use seed sentences tend to yield longer hypotheses and tend not to show a clear relationship between hypothesis--premise token overlap and label. \taskf\ tends to produce  shorter hypotheses.

\paragraph{Time Cost}

Annotators completed each of the five protocols at a similar rate, taking 3--4 minutes per prompt. This goes against our expectations that the longer premises in \taskb\ should substantially slow the annotation process, and that the pre-filled text in \taskc\ and \taskd\ should speed annotation. Since the relatively complex \taskf\ produces only two sentence pairs per prompt rather than three, it yields fewer examples per minute.

\paragraph{Label--Word Associations}

\begin{table}[t!]

\small

\centering
\begin{tabular}{llrr@{/}l}
\toprule
\textbf{Word}     & \textbf{Label} & \textbf{PMI} & \multicolumn{2}{c}{\textbf{Counts}} \\
\midrule
\midrule
\multicolumn{5}{c}{\textbf{\mnli\ 8.5k}}                                                    \\
\midrule
\textit{no}            & contradiction & 0.931 & 407       & 461   \\
\textit{any}           & contradiction & 0.809 & 169       & 208   \\
\textit{never}         & contradiction & 0.749 & 75        & 90    \\
\textit{nothing}       & contradiction & 0.721 & 43        & 47    \\

\midrule
\multicolumn{5}{c}{\textbf{\mnli\ Gov. 8.5k}}                                                     \\
\midrule
\textit{never}         & contradiction & 0.837 & 152       & 178   \\
\textit{no}            & contradiction & 0.828 & 342       & 426   \\
\textit{any}           & contradiction & 0.721 & 128       & 169   \\
\textit{nothing}       & contradiction & 0.712 & 56        & 66    \\

\midrule
\multicolumn{5}{c}{\textbf{\taska}}                                          \\
\midrule
\textit{never}         & contradiction & 0.935 & 231       & 255   \\
\textit{also}          & neutral       & 0.587 & 64        & 93    \\
\textit{any}           & contradiction & 0.585 & 46        & 64    \\
\textit{no}            & contradiction & 0.561 & 75        & 116   \\

\midrule
\multicolumn{5}{c}{\textbf{\taskb}}                                                          \\
\midrule
\textit{never}         & contradiction & 0.608 & 49        & 67    \\
\textit{than}          & neutral       & 0.526 & 95        & 156   \\
\textit{went}          & neutral       & 0.489 & 46        & 73    \\
\textit{lot}           & neutral       & 0.470 & 14        & 15    \\

\midrule
\multicolumn{5}{c}{\textbf{\taskc}}                                                          \\
\midrule
\textit{years}         & neutral       & 0.461 & 135       & 239   \\
\textit{ago}           & neutral       & 0.443 & 17        & 21    \\
\textit{eight}         & contradiction & 0.437 & 13        & 15    \\
\textit{refused}       & contradiction & 0.437 & 13        & 15    \\

\midrule
\multicolumn{5}{c}{\textbf{\taskd}}                                                          \\
\midrule
\textit{refused}       & contradiction & 0.565 & 24        & 28    \\
\textit{hardly}        & contradiction & 0.507 & 16        & 17    \\
\textit{later}         & neutral       & 0.482 & 48        & 77    \\
\textit{also}          & neutral       & 0.448 & 99        & 178   \\

\midrule\midrule
\multicolumn{5}{c}{\textbf{\mnli\ Gov. 8.5k (two-class)}}                                                          \\
\midrule
\textit{no}            & contradiction & 0.754 & 437       & 461   \\
\textit{any}           & contradiction & 0.689 & 193       & 208   \\
\textit{only}          & contradiction & 0.633 & 215       & 249   \\
\textit{never}         & contradiction & 0.625 & 86        & 90    \\

\midrule
\multicolumn{5}{c}{\textbf{\taskf~ (two-class)}}                                                      \\
\midrule
\textit{only}          & contradiction & 0.677 & 176       & 228   \\
\textit{never}         & contradiction & 0.635 & 73        & 90    \\
\textit{no}            & contradiction & 0.616 & 102       & 135   \\
\textit{not}           & contradiction & 0.571 & 156       & 226   \\

\bottomrule
\end{tabular}
\caption{The top four words most associated with specific labels in each dataset, sorted by the PMI between the word and the label. The \textit{counts} column shows how many of the instances of each word occur in hypotheses matching the specified label. We compare the two-class \taskf\ with a two-class version of MNLI Gov.}\label{tab:pmi}
\end{table}

Table \ref{tab:pmi} shows the four words in each dataset that are most predictive of example labels, using the smoothed PMI method of \citet{gururangan2018}. We also include results for two baselines: 8.5k-example samples from MNLI, and from MNLI's the \textit{government documents} single-genre section, which is meant to to be maximally comparable to the single-genre datasets we collect.

\taska\ shows similar associations to \mnli, but all four of our interventions reduce these  associations at least slightly. The use of seed sentences, especially in \taskc{}, largely eliminates the strong association between negation and contradiction seen in MNLI, and no new strong associations appear to take its place.

\section{Modeling Experiments}

We run three types of machine learning experiments: Sanity check experiments where we train and test on the NLI task---both in a standard setting and in a \textit{hypothesis-only} limited-input setting to measure relevant annotation artifacts---and our primary evaluation experiments in which we train models on NLI before evaluating them on other tasks through transfer learning.

These experiments generally compare models trained on ten NLI datasets: Each of the five 8.5k-example training sets introduced in this paper; the full 393k-example MNLI training set; the full 1.1m-example ANLI training set (which combines the SNLI training set, the MNLI training set, and the supplemental ANLI training examples);\footnote{In these runs, we use only the original ANLI validation set for evaluation and early stopping.}  8.5k-example samples from the MNLI training set and from the combined ANLI training set, meant to control for the size differences between these existing datasets and our baselines; and finally an 8.5k-example sample from the government section of the MNLI training set, meant to control (as much as possible) for the difference between our single-genre Wikipedia datasets and MNLI's relatively diverse text.

Our models are trained starting from pretrained RoBERTa (\textit{large} variant) or XLNet \citep[\textit{large}, \textit{cased};][]{yang2019xlnet}. RoBERTa was at or near the state of the art on most of our target tasks as of the launch of our experiments. XLNet is competitive with RoBERTa on most tasks, it offers a natural replication, and because of its substantially different design, it mitigates issues with evaluating ANLI that arise because ANLI was collected with a model-in-the-loop procedure using RoBERTa.

We run our experiments using \texttt{jiant} 1.2 \citep{wang2019jiant}, which implements the SuperGLUE tasks, MNLI, and ANLI, and in turn 
builds on \texttt{transformers} \citep{Wolf2019HuggingFacesTS}, AllenNLP \citep{Gardner2017AllenNLP}, and PyTorch \citep{paszke2017automatic}. To make it possible to train these large models on single consumer GPUs, we use small-batch ($b=4$) training and a maximum total sequence length of 128 word pieces.\footnote{We cut this to a slightly lower number on a few individual runs as needed to satisfy memory constraints. Note that this potentially limits the gains observable for \taskb{}, which has a longer mean premise length of 66.7 words.} We train for up to 2 epochs for the very large ReCoRD, 10 epochs for the very small CB, COPA, and WSC, and 4 epochs for the remaining tasks. Except where noted, all results reflect the median final performance from three random restarts of training.\footnote{Scripts implementing our experiments are available at \url{https://github.com/nyu-mll/jiant/tree/nli-data}.
}

\paragraph{Direct NLI Evaluations}

As a preliminary sanity check, Table \ref{tab:nli-results} shows the results of evaluating models trained in each of the settings described above on their own validation sets, on the MNLI validation set, and on the expert-constructed GLUE diagnostic set \citep{wang2018glue}. As NLI classifiers trained on \taskf\ cannot produce the \textit{neutral} labels used in MNLI, we evaluate them separately and compare them with two-class variants of the MNLI models.

Our \taska\ data yields a model that performs somewhat worse than a comparable MNLI Gov. 8.5k model, both on the full MNLI validation set and on the GLUE diagnostic set. 
This suggests, at least tentatively, that 
the new annotations are significantly less consistent with the MNLI labeling standard. This is disconcerting, but does not interfere with 
our key comparisons. Precise comparisons between MNLI and our new data on in-domain test sets are not possible, since only MNLI has in-domain evaluation data that has undergone substantial quality control.

The main conclusion we  draw from these results is that none of the first three interventions improve performance on the out-of-domain GLUE diagnostic set, suggesting that they do not help in the collection of high-quality training data that is consistent with the MNLI label definitions. We also observe that the newer ANLI data yields \textit{worse} performance than MNLI on the out-of-domain evaluation data when we control for dataset size. 

\begin{table}[t!] 
\centering \small

\begin{tabular}{lccc}
\toprule
\textbf{Training Data} & 
\multicolumn{1}{c}{\textbf{Self}} & \multicolumn{1}{c}{\textbf{MNLI}} &  \textbf{GLUE Diag.}  \\ 
\midrule

\taska & 84.8 & 81.5 & 40.5\\
\taskb & 78.3 & 78.2 & 31.7\\
\taskc & 82.9 & 79.8 & 35.5\\
\taskd & 82.5 & 82.6 & 33.9\\
\mnli 8.5k & 87.5 & 87.5 & 44.6\\
\mnli Gov8.5k & 87.7 & 85.4 & 40.7\\
\anli 8.5k & 35.7 & 85.6 & 39.8\\
\mnli & 90.4 & \textbf{90.4}  & 49.2\\
\anli & 61.5 & 90.1 & \textbf{49.7}\\
\midrule
\mnli~(two-class) & 94.0 & \textbf{94.0} & --\\
\mnli 8.5k (two-class) & 92.4 & 92.4 & --\\
\taskf & 91.6 & 80.6 & --\\
\bottomrule
\end{tabular}
\caption{NLI modeling experiments with RoBERTa, reporting results on the validation sets for MNLI and for the task used for training each model (Self), and the GLUE diagnostic set (shown as Matthews Corr.). We compare the two-class \taskf\ with a two-class version of MNLI.
}
\label{tab:nli-results}
\end{table}

\begin{table}[t!] 
\centering \small

\begin{tabular}{lrr@{~~~~~}rr}
    \toprule
\textbf{Training Data} & \textbf{Self} & \textbf{MNLI}  \\
\midrule

\taska & 57.9 & 52.2\\
\taskb & 48.3 & 47.0\\
\taskc & 40.4 & 39.4\\
\taskd & 45.1 & 50.7\\
\mnli 8.5k & 56.8 & 56.8\\
\mnli Gov8.5k & 63.7 & 53.9\\
\anli 8.5k & 34.3 & 54.4\\
\mnli & 62.0 & \textbf{62.0}\\
\anli & 53.2 & 61.6\\
\midrule
\mnli~(two-class) & 72.6 & \textbf{72.6}\\
\mnli 8.5k (two-class) & 62.4 & \textbf{62.4}\\
\taskf & 56.9 & 55.9\\
\bottomrule
\end{tabular}
\caption{Results from RoBERTa hypothesis-only NLI classifiers on the vaidation sets for MNLI and for the datasets used in training.
}
\label{tab:ho}
\end{table}

\begin{table*}[t!] 
\centering \small 

\setlength{\tabcolsep}{0.4em}
\begin{tabular}{lccc@{/}ccc@{/}cc@{/}cccc} 

\toprule
\textbf{Intermediate-} & \multicolumn{1}{c}{\textbf{Avg.}} & \multicolumn{1}{c}{\textbf{BoolQ}} & \multicolumn{2}{c}{\textbf{CB}} &  \multicolumn{1}{c}{\textbf{COPA}} & \multicolumn{2}{c}{\textbf{MultiRC}} & \multicolumn{2}{c}{\textbf{ReCoRD}} & \multicolumn{1}{c}{\textbf{RTE}} & \multicolumn{1}{c}{\textbf{WiC}} & \multicolumn{1}{c}{\textbf{WSC}}   \\

\textbf{Training Data} & \multicolumn{1}{c}{$\mu$ ($\sigma$)} & \multicolumn{1}{c}{Acc.} & \multicolumn{2}{c}{F1/Acc.} &  \multicolumn{1}{c}{Acc.} & \multicolumn{2}{c}{F1$_a$/EM} & \multicolumn{2}{c}{F1/EM} & \multicolumn{1}{c}{Acc.} & \multicolumn{1}{c}{Acc.} & \multicolumn{1}{c}{Acc.} \\

\midrule
\multicolumn{13}{c}{\bf RoBERTa (large)}\\
\midrule
None                                    & 67.3 (1.2)                                 & 84.3          & 83.1          & 89.3          & 90.0          & 70.0          & 27.3          & 86.5          & 85.9          & 85.2          & \textbf{71.9}            & 64.4          \\ 
\taska                   & \textbf{72.2} (0.1) & 84.4          & \textbf{97.4} & \textbf{96.4} & \textbf{94.0} & 71.9          & 33.3          & 86.1          & 85.5          & 88.4          & 70.8                     & \textbf{76.9} \\ 
\taskb                   & 70.3 (0.1)                           & 84.7          & \textbf{97.4} & \textbf{96.4} & 90.0          & 70.4          & 29.9          & \textbf{86.7} & \textbf{86.0} & 86.3          & 70.2                     & 67.3          \\ 
\taskc                   & 69.6 (0.6)                           & 83.0          & 92.3          & 92.9          & 89.0          & 71.2          & 31.2          & 86.4          & 85.7          & 85.6          & 71.0                     & 65.4          \\ 
\taskd                   & 70.3 (0.1)                           & 84.2          & 91.8          & 94.6          & 91.0          & 70.7          & 31.3          & 86.2          & 85.6          & 87.4          & 71.5                     & 68.3          \\ 
\taskf                   & 69.2 (0.0)                           & 84.1          & 93.1          & 94.6          & 87.0          & 71.4          & 29.5          & 84.8          & 84.1          & 84.5          & 71.5                     & 67.3          \\ 
\mnli 8.5k               & 71.0 (0.6)                           & 84.7          & 96.1          & 94.6          & 92.0          & 71.7          & 32.3          & 86.4          & 85.7          & 87.4          & 74.0                     & 68.3          \\ 
\mnli Gov8.5k            & 70.9 (0.5)                           & \textbf{84.8} & \textbf{97.4} & \textbf{96.4} & 92.0          & 71.4          & 32.0          & 86.2          & 85.6          & 86.3          & 71.6                     & 70.2          \\ 
\anli 8.5k               & 70.5 (0.3)                           & 84.7          & 96.1          & 94.6          & 89.0          & 71.6          & 31.8          & 85.7          & 85.0          & 85.9          & \textbf{71.9}            & 70.2          \\ 
\mnli                    & 70.0 (0.0)                                 & 85.3          & 89.0          & 92.9          & 88.0          & \textbf{72.2} & 35.4          & 84.7          & 84.1          & 89.2          & 71.8                     & 66.3          \\ 
\anli                    & 70.4 (0.9)                                 & 85.4          & 92.4          & 92.9          & 90.0          & 72.0          & 33.5          & 85.5          & 84.8          & \textbf{91.0} & 71.8                     & 66.3          \\ 
\midrule
\multicolumn{13}{c}{\bf XLNet (large cased)}\\
\midrule
None                                    & 62.7 (1.3)                                 & 82.0          & 83.1          & 89.3          & 76.0          & 69.9          & 26.8          & 80.9          & 80.1          & 69.0          & 65.2                     & 63.5          \\ 
\taska                   & 67.7 (0.0)                           & 83.1          & 90.5          & 92.9          & \textbf{89.0} & 70.5          & 28.2          & 78.2          & 77.4          & 85.9          & 68.7                     & 64.4          \\ 
\taskb                   & 67.3 (0.0)                           & 82.5          & \textbf{90.8} & \textbf{94.6} & 85.0          & 69.8          & 28.1          & 79.4          & 78.6          & 83.8          & 69.7                     & 64.4          \\ 
\taskc                   & 67.0 (0.4)                           & 82.8          & 82.8          & 91.1          & 83.0          & 69.8          & 28.6          & 79.3          & 78.5          & 85.2          & \textbf{70.2}            & \textbf{65.4} \\ 
\taskd                   & 67.2 (0.1)                           & 82.9          & 84.4          & 91.1          & 87.0          & 70.2          & 29.1          & 79.4          & 78.6          & 85.6          & 69.7                     & 63.5          \\ 
\taskf                   & 66.3 (0.6)                           & 83.0          & 82.5          & 89.3          & 83.0          & 69.8          & 28.3          & 80.2          & 79.5          & 85.9          & 68.2                     & 58.7          \\ 
\mnli 8.5k               & 67.6 (0.1)                           & 83.5          & 89.5          & 92.9          & 88.0          & 69.4          & 28.3          & 79.5          & 78.6          & 86.3          & 69.3                     & 62.5          \\ 
\mnli Gov8.5k            & 67.5 (0.3)                           & 82.5          & 89.5          & \textbf{94.6} & 85.0          & 70.0          & 28.1          & 79.8          & 79.0          & 87.4          & 68.7                     & 62.5          \\ 
\anli 8.5k               & 67.2 (0.3)                           & 83.4          & 86.3          & 91.1          & 83.0          & 69.3          & 28.9          & \textbf{81.2} & \textbf{80.4} & 85.9          & 70.1                     & 63.5          \\ 
\mnli                    & 67.7 (0.1)                                 & \textbf{84.0} & 85.5          & 91.1          & \textbf{89.0} & \textbf{71.5} & \textbf{31.0} & 79.1          & 78.3          & 87.7          & 68.5                     & 63.5          \\ 
\anli                    & \textbf{68.1} (0.4)       & 83.7          & 82.8          & 91.1          & 86.0          & 71.3          & 30.0          & 80.1          & 79.3          & \textbf{89.5} & 69.6                     & 66.3          \\ 
\bottomrule
\end{tabular}
\caption{Model performance on the SuperGLUE task validation sets. The Avg. column shows the overall SuperGLUE score---an average across the eight tasks
---as a mean and standard deviation across three restarts. 
}
\label{tab:benchmark}
\end{table*}

\paragraph{Hypothesis-Only Models}

To further investigate the degree to which our hypotheses contain artifacts that reveal their labels, Table \ref{tab:ho} shows results with single-input versions of our models trained on hypothesis-only versions of the datasets under study and evaluated on the datasets' validation sets.

Our first three interventions, especially \taskc, show much lower hypothesis-only  performance than \taska. This drop is much larger than the drop seen in our standard NLI experiments in the \textit{Self} column of Table \ref{tab:nli-results}. This indicates that these results cannot be explained away as a consequence of the lower label consistency of the evaluation sets for these three new datasets. This adds further evidence, alongside our PMI results, that these interventions reduce the presence of such artifacts. While we do not have a direct baseline for the two-class \taskf\ in this experiment, comparisons with MNLI 8.5k are consistent with the encouraging PMI results seen above.

\paragraph{Transfer Evaluations}

For our primary evaluation, we use the training sets from our datasets in STILTs-style intermediate training \citep{phang2018sentence}: We fine-tune a large pretrained model on our collected data using standard fine-tuning procedures, then fine-tune copies of the resulting model again on each of the target task datasets we use. We then measure the aggregate performance of the resulting models across those evaluation datasets.

We evaluate on the target tasks in the SuperGLUE benchmark \citep{wang2019superglue}: which consists of standardized splits and metrics for the question answering tasks 
BoolQ \citep{clark2019boolq}, MultiRC \citep{khashabi2018looking},  ReCoRD \citep{zhang2018record}; the entailment and reasoning tasks
CommitmentBank \citep[CB;][]{demarneffe:cb}, Choice of Plausible Alternatives \citep[COPA;][]{roemmele2011choice}, Recognizing Textual Entailment \citep[RTE;][]{dagan2006pascal,bar2006second,giampiccolo2007third,bentivogli2009fifth}, and 
the Winograd Schema Challenge \citep[WSC;][]{levesque2012winograd}; and  the word sense disambiguation task
WiC \citep{pilehvar2018wic}. 
These tasks were selected to be difficult for BERT but relatively easy for crowdworkers, and are meant to replace the largely-solved GLUE benchmark \citep{wang2018glue}. 

SuperGLUE does not include labeled test data, and does not allow for substantial ablation analyses on its test sets. Since we have no single final model whose performance we aim to show off, we do not use the test sets.
We train our WSC model in the standard way without adding data or modifying the format \citep[as in][]{kocijan2019surprisingly,liu2019roberta}. Without these modifications, few of our models exceed chance accuracy.

Results are shown in Table \ref{tab:benchmark}. Our first observation is that our overall data collection pipeline worked well for our purposes: Our \taska{} data yields models that transfer substantially better than the plain RoBERTa or XLNet baseline, and at least slightly better than 8.5k-example samples of MNLI, MNLI Government or ANLI. 
However, all four of our interventions yield \textit{worse} transfer performance than \taska{}. The variances across runs are small, and this pattern is consistent across both RoBERTa and XLNet, and across most individual target tasks.
We believe that this is a genuine negative result: At least under the broad experimental setting outlined here, we find that none of these four interventions is helpful for transfer learning. 

We chose to collect 8,500-example samples because of the prior observation that this approximate amount was sufficient to show clear results on transfer learning, and we reproduce that finding here: Both MNLI 8.5k and the \taska{} dataset yield large improvements over plain RoBERTa or XLNet through transfer learning. If any of our interventions were to be helpful in general, we would expect them to be harmless or helpful in our regime relative to \taska{}. This is not what we observe. 

We believe that this is the first study to evaluate ANLI as a pretraining task in transfer learning, and we observe that the large combined ANLI training set yields consistently better transfer than the original MNLI dataset. However, we observe (to our surprise) that this result reverses when we control for ANLI's larger size, with an 8.5k-example sample of MNLI yielding consistently better performance than an equivalently small sample of ANLI.

Our best overall result uses only 8.5k NLI training examples, suggesting either that this size is enough to maximize the gains available through NLI pretraining, or that the potential for models to forget skills learned in pretraining makes using larger intermediate datasets more challenging.

Finally, we replicate the finding from \citet{phang2018sentence}
that intermediate-task training with NLI data substantially reduces the variance across restarts seen in target task tuning.

\section{Conclusion}

Our chief results on transfer learning are conclusively negative: All four interventions yield substantially worse transfer performance than our base  MNLI data collection protocol. 
However, we also observe promising signs that all four of our interventions help to reduce the prevalence of artifacts in the generated hypotheses that reveal the label. While these interventions may be helpful for future \textit{evaluation} data, it appears that the type of creativity induced by our relatively open-ended \taska{}\ prompt works well for pretraining, and the resulting artifacts are a tolerable side-effect of that creativity.

The need and opportunity that motivated this work remains compelling: Human-annotated data like MNLI has already proven itself as a valuable tool in teaching machines general-purpose skills for language understanding, and discovering ways to more effectively build and use such data could further accelerate the field's already fast progress toward robust, general-purpose language understanding technologies. 

On another note, most available text corpora, including our Wikipedia source text and comparable past NLI datasets, contain evidence of social inequalities and stereotypes, which models can easily learn to reproduce \citep{wagner2015s,rudinger-etal-2017-social}. Our interventions  are not meant to address this, and are likely orthogonal. Bias mitigation in models and datasets remains a crucial direction for future work if systems based on datasets like the ones we study are to be widely deployed.

Beyond this:  Work on incentive structures and task design could facilitate the creation of crowdsourced datasets that are both creative and consistently labeled. Machine learning methods work on transfer learning could help to better understand and exploit the effects that drive the successes we have seen with NLI data so far. Finally, there remains room for further empirical work investigating the kinds of task definitions and data collection protocols most likely to yield training data that teaches models transferrable skills.
 
\section*{Acknowledgments}

We thank the annotators who spent time and effort on this project and the many members of the natural language processing community at Google who provided feedback.

\bibliography{emnlp-ijcnlp-2019}
\bibliographystyle{acl_natbib}

\end{document}